# Leveraging LLM for Automated Ontology Extraction and Knowledge Graph Generation

Mohammad Sadeq Abolhasani, Arizona State University

Rong Pan, PhD, Arizona State University

*SUMMARY & CONCLUSIONS*

Extracting relevant and structured knowledge from large, complex technical documents within the Reliability and Maintainability (RAM) domain is labor-intensive and prone to errors. Our work addresses this challenge by presenting OntoKGen, a Genuine pipeline for Ontology extraction and Knowledge Graph (KG) generation. OntoKGen leverages Large Language Models (LLMs) through an interactive user interface guided by our adaptive iterative Chain of Thought (CoT) algorithm to ensure that the ontology extraction process and, thus, KG generation align with user-specific requirements. Although KG generation follows a clear, structured path based on the confirmed ontology, there is no universally correct ontology as it is inherently based on the user's preferences. OntoKGen recommends an ontology grounded in best practices, minimizing user effort and providing valuable insights that may have been overlooked, all while giving the user complete control over the final ontology. Having generated the KG based on the confirmed ontology, OntoKGen enables seamless integration into schemeless, non-relational databases like Neo4j. This integration allows for flexible storage and retrieval of knowledge from diverse, unstructured sources, facilitating advanced querying, analysis, and decision-making. Moreover, the generated KG serves as a robust foundation for future integration into Retrieval-Augmented Generation (RAG) systems, offering enhanced capabilities for developing domain-specific intelligent applications.

## 1 INTRODUCTION

The complexity and volume of records and technical documents have been rapidly increasing across various engineering fields, including the semiconductor sector. These documents, essential for ensuring equipment's reliability and maintainability (RAM), often span hundreds of pages and encompass a broad array of topics. Engineers need to quickly access and locate specific information without sifting through entire sources and documents to find their desired information, where key details may be buried within extensive technical manuals. Beyond merely retrieving information, they also require tools to visualize data, uncover hidden relationships, and make inferences to support decision-making. This highlights the necessity for an efficient, automated system capable of extracting, organizing, and leveraging this knowledge for various analytical purposes, which is the primary motivation behind developing an LLM-assisted platform.

Transforming knowledge into a graspable form that is easy to transmit, and process is vital for effective operations. Knowledge Graphs (KGs) are pivotal in organizing and representing this knowledge [1]. KGs enable the seamless connection of vast amounts of heterogeneous information buried in engineering documents, offering a unified view that facilitates predictive maintenance, failure analysis, and decision-making. By linking different pieces of information, KGs enhance the ability to uncover hidden patterns and relationships, leading to improved operational efficiency and reduced downtime.

Ontologies provide the structural foundation for describing and structuring domain knowledge within KGs, representing the entities and relationships specific to the technical documents. However, constructing ontologies and KGs has traditionally required extensive collaborative and interdisciplinary efforts, often demanding significant time and expertise [2]. LLMs offer a promising solution by automating the extraction of ontologies and the generation of KGs as they encode rich world and domain-specific knowledge, enabling efficient information extraction through well-designed prompting techniques [3].

## 2 LITERATURE REVIEW

The expertise required in graph structures, web technologies, existing models, vocabularies, rule sets, logic, and best practices poses significant challenges in constructing meaningful KGs. Meyer et al. (2024) investigated the potential of ChatGPT to address these challenges, demonstrating that ChatGPT can automatically generate KG, though validation and refinement were often required to ensure accuracy. Tahsin et al. (2024) used LLMs to tackle the problem of formalizing knowledge from unstructured maintenance work orders. They proposed a framework for creating semantic KGs using a Simple Knowledge Organization System (SKOS) thesaurus extended with a fine-tuned LLM. Kommineni et al. (2024)

explored the semi-automatic construction of KGs using open-source LLMs. Their approach involves focusing on competency questions to develop an ontology and generating KGs with minimal human involvement. Carta et al. (2023) introduced a strategy for KG generation based on zero-shot prompting with LLMs.

Despite these advancements, there remain gaps in handling broader and more complex topics, such as those in lengthy and intricate engineering documents. Much of the prior work has focused on narrow and fixed domains, like failure detection, where all instances follow narrow and predefined concepts and relationships in the ontology. In contrast, engineering documents and standards span a much wider array of topics.

Our work aims to bridge these gaps with an interactive platform that employs an adaptive iterative CoT algorithm. The system operates automatically, guided by built-in algorithms for ontology extraction and KG generation while allowing user interaction to refine the process at any stage. This combination significantly enhances the efficiency and accuracy of knowledge extraction and representation.

## 3 METHODOLOGY

### 3.1 Overview of the Proposed System

OntoKGen is designed to generate KGs that are tailored specifically for the RAM domain. OntoKGen integrates OpenAI's API to harness the power of LLMs to automate the extraction of domain-specific ontologies and generate KGs based on user inputs without requiring local computational resources. At the core of the OntoKGen (as shown in Figure 1) is an interactive conversational interface, which engages the user in a dialogue to gather the necessary information for ontology extraction and, subsequently, KG generation.

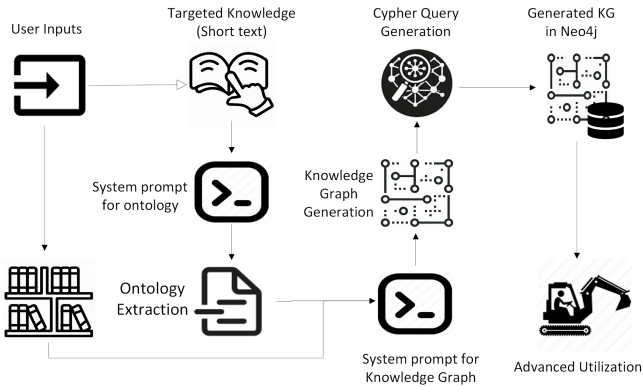

*Figure 1 Overview of OntoKGen's pipeline*

The system prompts users to provide either a predefined ontology or a piece of text (targeted knowledge) that focuses on the specific area or aspect of knowledge they are interested in. This targeted knowledge helps define the ontology, including relevant concepts (nodes), relationships (edges), and attributes (properties) pertinent to the RAM domain. Once the ontology is confirmed, the user is asked to provide comprehensive texts related to the engineering field of interest. OntoKGen then meticulously identifies relevant instances of the concepts and relationships from these texts to generate a detailed and accurate KG based on the confirmed ontology.

### 3.2 Chain of Thought (CoT) Prompting

Wei et al. (2023) introduced the concept of chain-of-thought prompting as a strategy that enhances the reasoning capabilities of language models by breaking down complex tasks into smaller, manageable steps. This approach enables the model to solve intricate problems more effectively by generating intermediate steps that lead to the final solution.

Conventional CoT prompting often involves adding brief instructions like "Let's think step by step" at the end of a prompt. This simple addition encourages the language model to record its reasoning process sequentially, which helps reduce errors and hallucinations.

Our approach goes further by not only asking the LLM to perform tasks step-by-step but also clearly defining each step. We incorporate our adaptive iterative CoT algorithm to ensure that the LLM executes tasks with consistency and precision. This methodology builds on the strengths of CoT prompting while adding rigor through explicitly defined steps, significantly improving the reliability and accuracy of the KG generation.

### 3.3 Ontology Extraction

Ontology extraction is a key component of OntoKGen, providing a structured framework that directs the knowledge graph (KG) generation process. An ontology defines concepts (nodes), relationships (edges), and the attributes (properties) of these nodes. Without an ontology, the KG would lack the necessary structure and coherence, making it difficult to extract meaningful insights and perform reliable analyses. Specifying the concepts and relationships in advance helps avoid ambiguities and ensures the integrity of the final KG.

---

**Algorithm 1** Ontology Extraction Instructions
---
**Require:** Targeted knowledge text $T_k$
**Ensure:** Extracted Ontology $O$
    **Step 1: Concept Identification**
1: Identify broader hierarchical concepts and categorize instances.
2: Find broader categories or abstract them if not present.
3: Use Example-Based Generalization to identify categories.
4: Present the initial concept list.
    Format: *Category Name: [Two instances as examples]*.
    **Step 2: Concept Confirmation**
5: Present the list of concepts and confirm with user interaction.
    **Step 3: Relationship Identification**
6: Identify relationships between confirmed concepts.
7: Provide the list of identified relationships.
    Format: *Relationship Name: [Concept1 → Concept2]*.
    **Step 4: Relationship Confirmation**
8: Confirm relationships with user interaction.
9: Present the complete ontology for final confirmation.
    **Step 5: Concepts' Properties**
10: Review the targeted knowledge text for additional attributes.
11: Present the list of identified properties.
    Format: *Property Name: [Concept Name → Property Name]*.
12: Confirm properties with user interaction.
13: Provide the complete confirmed ontology.

*Figure 2 Ontology Extraction Algorithm*

As shown in Figure 2, our adaptive ontology extraction algorithm involves identifying and confirming concepts, relationships, and properties through a series of structured steps that include user interaction and validation. This ensures a comprehensive and accurate ontology tailored to the users' needs, forming a robust foundation for KG generation.

### 3.4 Knowledge Graph Generation

Compared to the ontology extraction, which requires significant user interaction, the KG construction is more automated, as the ontology now serves as the structural blueprint for the KG generation and reflects all user-specific requirements. Our adaptive iterative CoT algorithm includes comprehensive steps and considerations to reduce the need for user intervention while allowing for necessary adjustments.

Once the ontology is confirmed, OntoKGen moves forward with the KG generation based on the algorithm illustrated in Figure 3. The ontology serves as a blueprint, ensuring the KG is generated with a clear and consistent representation of domain-specific knowledge. This process is crucial for creating a coherent and structured KG that accurately reflects the information extracted from the text.

---

**Algorithm 2** Knowledge Graph Generation Instructions
**Require:** Confirmed Ontology $O_f$, Comprehensive text $T_c$
**Ensure:** Generated Knowledge Graph $KG$

    **Phase 1: Creation**
    **Step 1: Identify Concepts**
1: **for** each concept $c$ in $O_f$ **do**
2:    Extract instances of $c$ from $T_c$
3:    Add nodes for each instance to $KG$ with properties Id and Name
4: **end for**
    **Step 2: Identify Relationships**
5: **for** each relationship $r$ in $O_f$ **do**
6:    Extract instances of $r$ from $T_c$
7:    Add edges between corresponding nodes in $KG$
8: **end for**
    **Step 3: Identify Properties**
9: **for** each property $p$ in $O_f$ **do**
10:   Extract instances of $p$ from $T_c$
11:   Add properties to corresponding nodes in $KG$
12: **end for**
    **Phase 2: Review**
    **Step 4: Review the whole text one more time**
13: Ensure all instances and relationships are accurately added to the $KG$.
    **Step 5: Review the whole Graph one more time**
14: Make sure all nodes, relationships, and properties are correctly represented.
15: Add any missing relationships to ensure a fully connected graph

*Figure 3 Knowledge Graph Generation Algorithm*

---

### 3.5 User Interface

The user interface is a critical component of OntoKGen, designed to facilitate smooth and efficient interaction between the user and the LLM. It guides users through the entire process of ontology extraction and KG generation, designed based on the best practices rooted in manual ontology extraction and knowledge discovery.

Through this interface, users can iteratively refine both the ontology and KG. This continuous engagement throughout the process keeps users involved in every step and significantly minimizes rework. The LLM uses the entire conversation history, drawing on both past and current user inputs, to maintain context, reduce errors and hallucinations, and improve the overall quality of the final KG.

We also leverage LLM to assist users in generating useful queries during the process. Users may need help remembering specific definitions, identifying concepts, or even some quick suggestions, and the LLM can provide valuable assistance in these areas. This collaborative environment allows users to benefit from the LLM's general knowledge and reasoning capabilities. Our user interface also maintains contextual awareness throughout the process. Even if off-topic questions arise from the user (Figure 4), OntoKGen is designed to address them briefly and then guide the user back to the current stage of the KG generation to ensure that the user remains on track.

### 3.6 Cypher Query Generation

Once the KG generation is complete, upon user confirmation, OntoKGen facilitates the transition of the KG to the Neo4j, a non-relational database with a flexible schema, using Cypher Query. By utilizing the MERGE command, the system ensures that nodes, relationships, and properties are created simultaneously. This prevents issues such as trying to create relationships for non-existent nodes and avoids duplicating nodes. This automated transition extends database utilization, which enables advanced querying, visualization, and inference via the generated KG within the Neo4j.

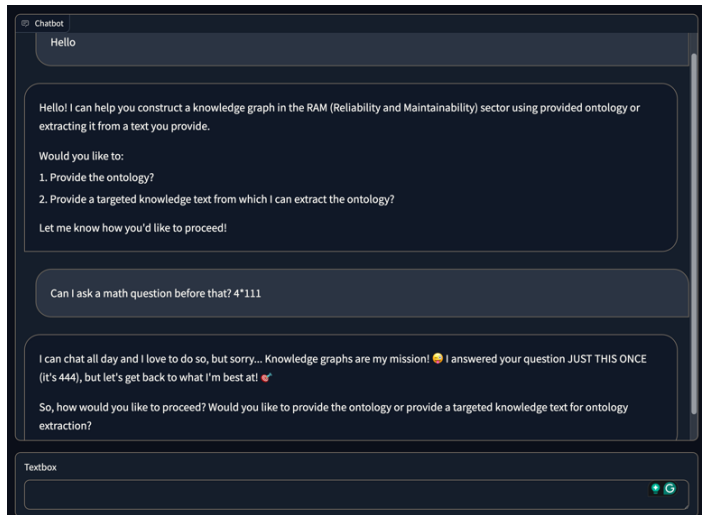

*Figure 4 OntoKGen's User Interface guided by the algorithms*

## 4 CASE STUDY

This case study demonstrates the implementation of OntoKGen within the RAM domain. It showcases the effectiveness and flexibility of OntoKGen in extracting and structuring domain-specific knowledge.

### 4.1 Problem Definition

The main challenge addressed in this case study involves the extraction of relevant and structured knowledge from large, complex technical documents within the RAM domain, specifically focusing on semiconductor manufacturing equipment. Traditional knowledge extraction methods are both labor-intensive and error-prone, often resulting in incomplete or inconsistent knowledge representations. Additionally, users must manually sift through lengthy and highly technical documents to locate specific information, which is time-consuming and difficult.

OntoKGen addresses these issues by automating this process, reducing the workload on domain experts while ensuring high accuracy and consistency in the resulting KG. The system also provides interactive, user-driven refinements at each step to meet specific user requirements.

### 4.2 *Data Collection*

For this case study, the document titled *Background Statement for Semiconductor Draft Document 6578* was used as the primary source for comprehensive text. Due to its length and technical complexity, manually extracting relevant information and constructing a KG from this document would be a challenging task. The document contains text, tables, and graphs with the primary focus on the textual content, which formed the basis for ontology extraction and KG generation.

Initially, users interact with OntoKGen by providing targeted knowledge that specifies their area of interest within the RAM domain. The targeted knowledge can range from a straightforward query, such as 'I want to know about equipment states' to more detailed one or even a part of the comprehensive text. For instance, in this case study, the targeted knowledge included definitions of concepts like "Productive State (PRD)" and associated activities and sub-states. OntoKGen used this input to structure the ontology and then generate a KG by reviewing the comprehensive text for relevant instances that align with the confirmed ontology.

### 4.3 *Ontology Extraction Process*

The ontology extraction process in this case study demonstrates the interactive and iterative nature of OntoKGen. The process begins by prompting the user to provide either a predefined ontology or a short text specifying the targeted knowledge. Based on the user's input, OntoKGen suggests additional concepts to ensure broader coverage. For example, in this case study, the user initially provided the concept of "Productive State (PRD)" along with related activities. OntoKGen proposed expanding the scope by including higher-level concepts such as "Equipment System," demonstrating its ability to refine and enhance the ontology for greater consistency and coverage.

Upon user confirmation of the identified concepts, OntoKGen then identified relationships between the confirmed concepts, such as "Has State," "Has Metric," and "Has Activity." User feedback is integrated to refine these relationships, including hierarchical associations within equipment states or activities tied to sub-states. OntoKGen then, carried out relationship confirmation by presenting the identified relationships to the user for final approval.

Finally, the system reviews the targeted knowledge to identify additional attributes for nodes beyond id and name, such as a "brief explanation," and suggests these enhancements to the user to improve clarity. Upon user approval on the properties, OntoKGen presents the complete ontology for final approval, ensuring that the final ontology is accurate, comprehensive, and tailored to the user's specific needs, establishing a solid foundation for the subsequent KG generation phase.

The final confirmed ontology for this case study comprised the following key elements:

**Concepts**: Equipment System, Equipment States (e.g., Productive State, Scheduled Downtime State), Sub-States (e.g., SDT preventive maintenance, SDT setup), Activities (e.g., Regular production, Rework), Metrics (e.g., Equipment-Dependent Metrics)

**Relationships**: Has State, Has Sub-State, Has Activity, and Has Metric.

**Properties**: Mandatory properties for each node (name, id), with user-specified attributes "brief explanation".

### 4.4 *Knowledge Graph Generation Process*

The KG generation process begins once the final ontology is confirmed by the user. OntoKGen then requests the comprehensive text, in this case "Background Statement for Semiconductor Draft Document 6578". Following the adaptive iterative CoT algorithm, OntoKGen carries out KG generation in two main phases: Creation and Review.

**Phase 1**: Creation

The system first reviews the confirmed ontology and begins by identifying relevant concepts and relationships from the comprehensive text

**Step 1**: Identify Concepts

OntoKGen starts by examining the comprehensive text to find instances that match the first concept in the ontology, such as "Equipment System." A node is created for each identified instance, with properties including an ID, name, and a brief explanation. For example, a node for "Equipment System" may have the following properties:

- **Node:** Equipment System
- **Properties:** {id: "equipmentSystem1", name: "Equipment System", briefExplanation: "Central node containing all equipment states, activities, and metrics."}

The system then identifies the next concept in the ontology, such as "Equipment States," and repeats the process iteratively until all concepts are covered.

**Step 2**: Identify Relationships

OntoKGen then identifies the first relationship in the confirmed ontology, carefully reviews the text to find instances that align with this relationship and creates relationships between the corresponding nodes. For example, the relationship "Has State" is made between "Equipment System" and "Productive State (PRD)" based on the information in the text. OntoKGen then identifies the next relationship in the ontology and repeats the process until all relationships are covered.

**Step 3**: Identify Properties

The system then reviews the confirmed ontology to identify properties for each concept, adding properties to the corresponding nodes. Then, the process is repeated until all properties are covered. This ensured that all relevant properties were accurately represented in the KG.

**Phase 2**: Review

**Step 4**: Review the Text

OntoKGen reviews the entire comprehensive text again to ensure all instances and relationships mentioned in the text, which align with the ontology, are represented in the KG. Any missing nodes or relationships are added as required.

**Step 5**: Review the Graph

Finally, the system performs a final review of the entire constructed KG to ensure all nodes, relationships, and properties are correctly represented according to the confirmed ontology and the text. It checks for unconnected nodes and adds necessary relationships to ensure a fully connected graph.

Throughout this process, the system strictly adheres to several additional predefined rules, such as using human-readable names, abbreviations, and identifiers, to maintain uniformity and set a solid foundation for further querying and analysis.

### 4.5 Cypher Query Generation

After generating the KG based on the confirmed ontology and comprehensive text, the final step in the pipeline prompts the user to specify whether Cypher queries are needed to import the generated KG into Neo4j.

Upon approval, OntoKGen generates the necessary Cypher queries using the MERGE command to create nodes and relationships simultaneously to prevent issues such as duplicate nodes or attempts to create relationships for non-existent nodes. Then, the complete Cypher commands are provided for execution in Neo4j.

### 4.6 Results and Discussion

The generated KG in Neo4j demonstrates OntoKGen's ability to transform comprehensive technical documents into structured, actionable knowledge. Figures 5 and 6 provide visualizations of the generated KG. The KG includes various nodes, relationships, and properties, and the following are some key examples from the KG highlighting essential benefits.

Central Node

The *Equipment System* node serves as the central hub, connecting all other nodes in the graph. This centralization enables the interconnection of all relevant data points, providing a holistic view of the equipment's states, activities, and metrics.

Interconnected Equipment States

Relationships between the Equipment System and various Equipment States (e.g., Productive State) allow users to easily navigate between different nodes, in this case states, and understand their connections and dependencies.

Detailed Sub-States and Activities

Linking sub-states and activities to their respective equipment states provides a detailed breakdown of the equipment's operational context. This granularity allows users to pinpoint specific activities and sub-states that impact equipment performance, facilitating targeted analysis and improvements.

Comprehensive Metrics

Metrics nodes connected to the Equipment System enable users to evaluate performance metrics in context. This integration supports comprehensive performance analysis, helping to pinpoint areas for potential improvement.

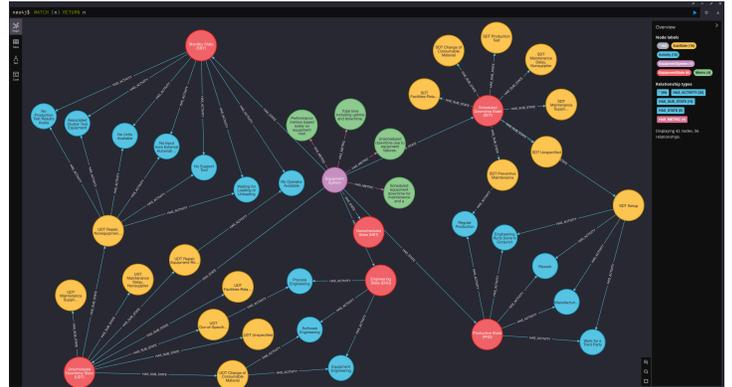

*Figure 5 Generated Knowledge Graph in Neo4j*

Accuracy and Completeness

The generated KG accurately reflects the information provided in the comprehensive text. Every concept, sub-state, activity, and metric is correctly represented, showcasing OntoKGen's effectiveness in extracting and structuring domain-specific knowledge.

Efficiency

The automated process significantly reduces the manual effort required to construct a KG from lengthy and complex documents. OntoKGen's capability to dynamically update and refine the ontology based on user input ensures the final KG is comprehensive and tailored to the user's needs.

User Interaction

While the KG generation phase required minimal user interaction, the initial ontology extraction phase ensured that all user-specific requirements were incorporated. This balance between automation and user involvement enhances both the efficiency and accuracy of the final KG.

Integration into Neo4J

The generated KG in Neo4j serves as a powerful tool for exploring and analyzing RAM-related data. OntoKGen's ability to transform extensive technical documents into structured knowledge not only saves time and reduces errors but also provides a comprehensive, user-tailored knowledge graph that supports advanced data operations and insights.

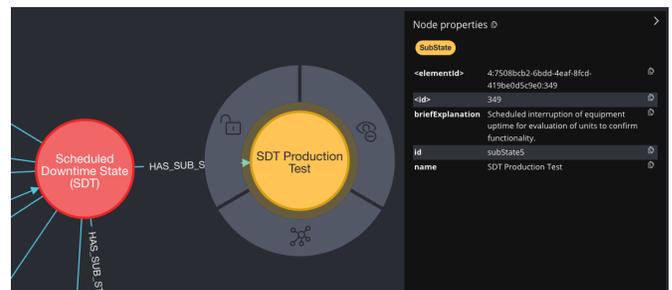

*Figure 6 Extracted node's properties*

## 5 CONCLUSION AND FUTURE WORKS

In this paper, we demonstrated how OntoKGen facilitates ontology extraction and KG generation within the Reliability and Maintainability domain, specifically applied to semiconductor equipment. OntoKGen's interactive approach, combined with our adaptive iterative CoT algorithm, ensures that the ontology and generated KG are tailored to user-specific requirements, resulting in a structured and meaningful knowledge representation.

By automating the extraction process and incorporating user interaction at critical stages, OntoKGen significantly reduces manual effort and minimizes errors. The generated KGs provide a robust framework for advanced querying, analysis, and decision-making. Additionally, these KGs enhance inference capabilities by revealing relationships that might otherwise be difficult to identify manually. Furthermore, KGs constructed in various domains can serve as valuable sources for Retrieval-Augmented Generation (RAG) applications, paving the way for creating more intelligent domain-specific AI systems in the future.

For this paper, we employed a human-in-the-loop approach to evaluate the performance of OntoKGen. Although we have set the foundation for competency question answering to evaluate the method's accuracy, conducting this quantitative analysis and presenting those results are beyond the scope and limitations of this paper. Therefore, they will be addressed in future work.

Our Future work will focus on leveraging the generated KGs as sources in Retrieval-Augmented Generation (RAG) systems, enhancing the development of more intelligent and domain-specific AI solutions. Additionally, we aim to implement live data manipulation directly from the interactive interface, enabling users to update and modify the KG dynamically. These advancements will significantly enhance the intracavity and usability of the system, making it a more powerful tool for knowledge management and decision-making.

## BIOGRAPHIES

Mohammad Sadeq Abolhasani
School of Computing and Augmented Intelligence
Arizona State University
699 S. Mill Avenue, Suite 225
Tempe, AZ 85281 USA

e-mail: mabolhas@asu.edu

Mohammad Sadeq Abolhasani is a PhD student in Data Science, Engineering, and Analytics at School of Computing and Augmented Intelligence, Arizona State University. He holds a bachelor's degree in industrial engineering and earned his master's degree in industrial management from the University of Tehran in 2021. His research leverages Large Language Models and Retrieval-Augmented Generation to advance knowledge extraction, representation, and inference in complex technical domains. Mohammad is particularly interested in integrating AI-driven techniques with graphs to enhance information retrieval, representation, and decision-making processes.

Rong Pan, PhD
School of Computing and Augmented Intelligence
Arizona State University
699 S. Mill Avenue, Suite 225
Tempe, AZ 85281 USA

e-mail: rong.pan@asu.edu

Rong Pan is a Professor of Industrial Engineering at Arizona State University. He received his doctoral degree in Industrial Engineering from Penn State University in 2002. His research interests include failure time data analysis, system reliability, design of experiments, multivariate statistical process control, time series analysis, and computational Bayesian method.